\documentclass[10pt,twocolumn,letterpaper]{article}

\usepackage{iccv}
\usepackage{times}
\usepackage{epsfig}
\usepackage{graphicx}
\usepackage{amsmath}
\usepackage{amssymb}
\usepackage{booktabs}
\usepackage{svg}
\usepackage{nicefrac}

\newcommand{\Hquad}{\hspace{1em}} 

\usepackage{hyperref}
\hypersetup{pagebackref=true,breaklinks=true,letterpaper=true,colorlinks,bookmarks=false}

\iccvfinalcopy

\ificcvfinal\fi

\newcommand\blfootnote[1]{%
  \begingroup
  \renewcommand\thefootnote{}\footnote{#1}%
  \addtocounter{footnote}{-1}%
  \endgroup
}

\begin{document}

\title{Audio-Enhanced Text-to-Video Retrieval using Text-Conditioned Feature Alignment}

\author{Sarah Ibrahimi$^1$\thanks{Work done while interning at Amazon Prime Video.} \Hquad Xiaohang Sun$^2$ \Hquad Pichao Wang$^2$ \Hquad Amanmeet Garg$^2$ \\ \Hquad Ashutosh Sanan$^2$ \Hquad Mohamed Omar$^2$\\
$^1$ University of Amsterdam \quad $^2$ Amazon Prime Video\\
{\tt\small s.ibrahimi@uva.nl, \{sunking, wpichao, amanmega, ashsanan, omarmk\}@amazon.com}
}

\maketitle

\ificcvfinal\thispagestyle{empty}\fi

\begin{abstract}
Text-to-video retrieval systems have recently made significant progress by utilizing pre-trained models trained on large-scale image-text pairs. However, most of the latest methods primarily focus on the video modality while disregarding the audio signal for this task. Nevertheless, a recent advancement by E\textsc{clip}SE has improved long-range text-to-video retrieval by developing an audiovisual video representation. Nonetheless, the objective of the text-to-video retrieval task is to capture the complementary audio and video information that is pertinent to the text query rather than simply achieving better audio and video alignment. To address this issue, we introduce TEFAL, a \textbf{TE}xt-conditioned \textbf{F}eature \textbf{AL}ignment method that produces both audio and video representations conditioned on the text query. Instead of using only an audiovisual attention block, which could suppress the audio information relevant to the text query, our approach employs two independent cross-modal attention blocks that enable the text to attend to the audio and video representations separately. Our proposed method's efficacy is demonstrated on four benchmark datasets that include audio: MSR-VTT, LSMDC, VATEX, and Charades, and achieves better than state-of-the-art performance consistently across the four datasets. This is attributed to the additional text-query-conditioned audio representation and the complementary information it adds to the text-query-conditioned video representation. \blfootnote{Accepted at ICCV 2023 (oral)}
\end{abstract}

\section{Introduction}
\label{sec:intro}

The emergence of online streaming services has led to an enormous and rapidly growing collection of multimedia assets comprising video and audio. Retrieving semantically similar content in these assets is crucial for finding information of interest, making it an important aspect of major streaming platforms. Text-to-video retrieval is a common approach to achieve this goal, whereby video content that best matches a textual description is searched for by learning a joint latent space for text and video representations. This space allows the text modality input to be matched with the video modality, enabling the closest videos to be found through a distance metric.

\begin{figure}[t]
  \centering
  \includegraphics[width=1.0\linewidth]{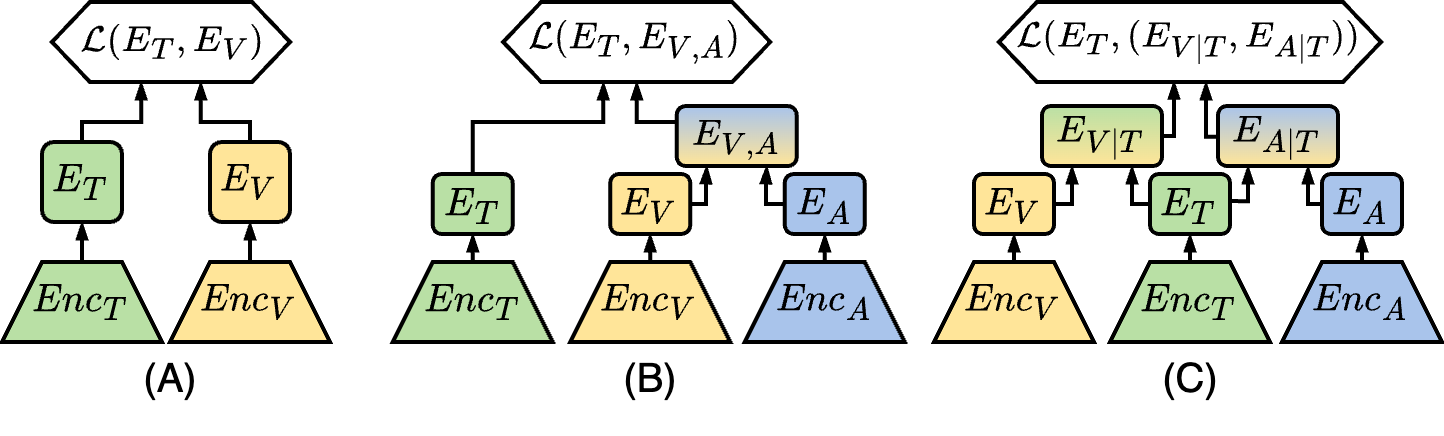}
   \caption{Comparison of our method with the current methods, text-to-video retrieval with (A) video only, (B) audio-video fusion and (C) proposed text-conditioned audio-video alignment (\textbf{\textit{TEFAL}}).}
   \label{fig:fig1}
\end{figure}

The rise of large-scale transformer models of vision and language has led to the development of many multimodal transformer-based architectures that are commonly evaluated on the text-to-video retrieval task. These architectures can either be pre-trained on large-scale multimodal datasets from scratch or use existing pre-trained models, such as CLIP~\cite{clip}, as starting points or frozen backbones. One of the earliest CLIP-based model architectures, CLIP4Clip \cite{Luo2022CLIP4ClipAE}, shows a significant improvement in performance compared to previous state-of-the-art methods \cite{Bain21frozen, lei2021clipbert, li2020hero} on common text-to-video retrieval benchmarks. By utilizing only a few video frames per video and a simple technique to aggregate the frame embeddings for each video, CLIP4Clip demonstrated the utility of a 2D vision transformer to outperform model architectures using 3D videos as input. Since then, many other works have been built upon this baseline approach with novel ways of cross-attention \cite{gorti2022xpool}, pretext tasks \cite{bridgeformer}, prompting \cite{alpro}, and other architectural modifications. Recently, significant improvements have been obtained by incorporating post-processing techniques, such as Querybank Normalization \cite{qbnorm}.

\begin{figure}[t]
   \centering
   \includegraphics[width=\linewidth]{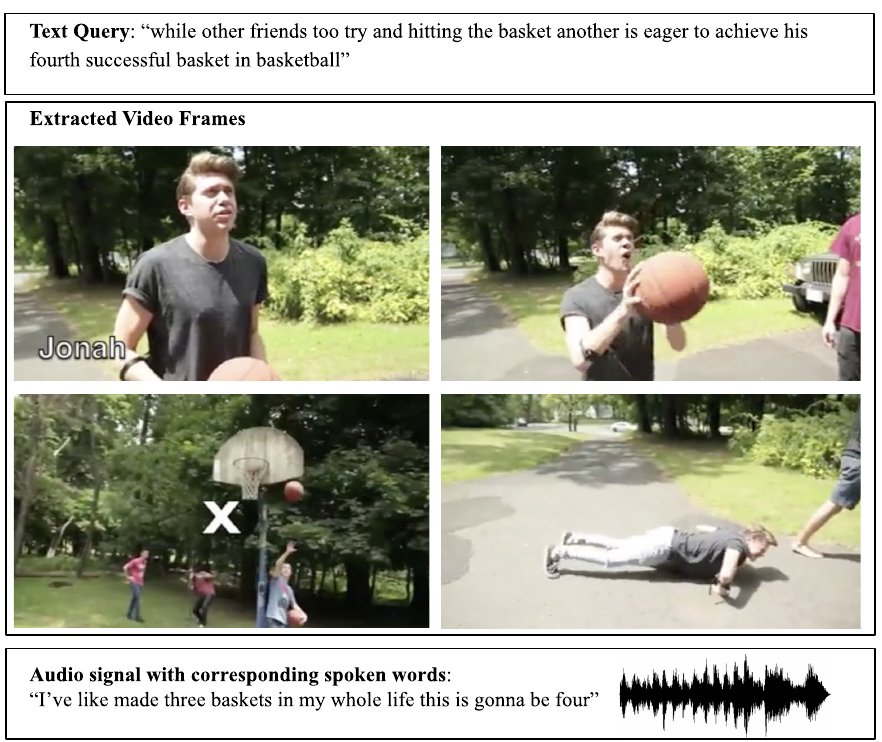}
   \caption{The audio signal can play a significant role in text-to-video retrieval. For instance, consider a textual description that mentions the fourth successful basket, which cannot be retrieved from the visual information alone. However, the spoken words in the video also contain the number four and can be matched to the query. Additionally, sounds of the basketball can further confirm the sport being played.}
   \label{fig:figw}
  \end{figure}

Most of the existing text-to-video retrieval models only consider the correspondence between text and video, as visualized in Figure~\ref{fig:fig1}(A), despite the fact that most videos also contain an audio track. As shown in Figure \ref{fig:figw}, audio can be related to actions, events, objects, and words in the caption, indicating that the audio signal could be beneficial for the task of text-to-video retrieval. Although the use of other modalities for text-to-video retrieval, including audio, has been studied in \cite{everythingatonce}, their results showed no improvement compared to the video-only setting. Recently, E\textsc{clip}SE \cite{ECLIPSE_ECCV22} successfully accounts for audio in the video representation by aligning audio and video using cross-modality attention, demonstrating its benefits for long-range text-to-video retrieval.

However, solely embedding the audio features for better video representations may not optimize the objective of text-to-video retrieval, which is to capture complementary audio and video information relevant to the text query. Videos can contain sounds that are not strongly correlated with the visual content, such as a gunshot without a visible gun in the video, as well as other types of off-screen sounds. If audio and video are processed jointly, important audio cues that are relevant to the text query but not visible in the video might be suppressed. To address this issue, we propose TEFAL, a framework that generates text-conditioned representation of both video and audio to achieve effective audio-enhanced text-to-video retrieval, as depicted in Figure~\ref{fig:fig1}. 

We use the CLIP~\cite{clip} model as the video and text feature extractor and the AST~\cite{gong21b_interspeech} model as the audio feature extractor in our approach. The text feature serves as a crucial link between the video and audio representations, acting as the query in the calculation for cross-attention. Meanwhile, the video and audio features are used for key and value computations. To simplify matters, the two aligned feature types are combined to produce the final audio-enhanced video representation. We conducted extensive experiments on various datasets that include audio, such as MSR-VTT~\cite{msrvtt}, LSMDC~\cite{lsmdc}, VATEX~\cite{vatex}, and Charades~\cite{charades}. Our proposed audio-enhanced text-conditioned feature alignment method consistently outperforms existing methods. Specifically, our method improves the Recall@1 by over 4\% compared to E\textsc{clip}SE on the MSR-VTT dataset. 

Our key contributions are summarised as follows:
\begin{itemize}
    \item We propose a text-conditioned feature alignment approach for audio-enhanced text-to-video retrieval. We are the first to do so and we explain why this approach is more suitable for this task than audiovisual alignment. To achieve this, we utilise two independent cross-modal attention blocks for the text to attend to the audio and video representations.
    \item  We conducted extensive experiments on several benchmark datasets that include audio, namely MSR-VTT, LSMDC, VATEX, and Charades. Our results demonstrate state-of-the-art performance in text-to-video retrieval when compared with the best previously published results.
\end{itemize}

\section{Related Work}
Our proposed method is closely related to text-to-video retrieval, multimodal video learning and audio-based multimodal learning. In the following we go over some of the main
works in these three directions and more comprehensive works are refereed to the survey papers~\cite{baltruvsaitis2018multimodal,bayoudh2022survey}. 

\subsection{Text-to-Video Retrieval}
The text-to-video retrieval task has gained significant attention in recent years \cite{Bain21frozen,croitoru2021teachtext,dong2021dual, ging2020coot,hu2022lightweight, lei2021clipbert,liu2022animating,liu2022ts2net,xclip,everythingatonce,wang2021t2vlad,zhang2018cross,zhu2020actbert}, where the goal is to retrieve relevant videos by a text query from a database of video clips. Prevailing works usually leverage model architectures pre-trained on large-scale text-to-video or text-image datasets \cite{Bain21frozen, lei2021clipbert, zhang2018cross, zhu2020actbert}. Upon the release of the CLIP model \cite{clip}, which consists of strong image and text backbones pre-trained on 400M image-text pairs, sparse sampling approaches started to gain popularity and achieved high performance with the release of CLIP4Clip \cite{Luo2022CLIP4ClipAE}. 
Since then, many text-to-video retrieval methods have taken CLIP and GPT~\cite{brown2020language, bridgeformer, alpro, xclip, radford2018improving,radford2019language} backbones as main components and improved by introducing \eg new cross-modal fusion \cite{gorti2022xpool} and token selection \cite{liu2022ts2net}. Our method also takes advantages of the pre-trained text-video models for robust feature extractions. 

\subsection{Multimodal Video Learning}
Multimodal video models focus on a wide range of tasks, such as visual commonsense reasoning, visual question answering, activity recognition and text-to-video retrieval~\cite{xu2022multimodal}. Li \textit{et al.} \cite{li2020hero} proposed to learn a hierarchical structure where the local context of a video frame is encoded first by a cross-modal transformer, followed by a temporal transformer to learn global video context embeddings. Contrary to works that use dense sampling of video frames and 3D features, ClipBERT~\cite{lei2021clipbert} introduced a pipeline that uses sparse sampling of video frames and simple 2D visual architectures during training. 

\subsection{Audio in Multimodal Learning}
Including the audio modality has been a topic of research in multimodal works. Merlot Reserve \cite{merlot_reserve} included the audio modality in large scale pre-training and designed a new contrastive mask training task where both text and audio are masked out. The Video-Audio-Text Transformer (VATT) \cite{vatt} is a convolution-free transformer architecture that can process multiple modalities including audio. Multimodal Versatile Networks \cite{MMVN} presents a self-supervised multimodal learning strategy. For text-to-video retrieval, \cite{miech18learning, collab_experts, wang2021t2vlad} integrated audio in the pipeline by using NetVLAD as an approach to aggregate audio features. MEE \cite{miech18learning} uses a scalar product between the text and audio feature and CE \cite{collab_experts} uses a MLP to model the pairwise relation between text and audio. T2VLAD \cite{wang2021t2vlad} uses a global-local alignment approach where clustering is used for the local alignment without explicitly conditioning on the text. 

More recently, Shvetsova \textit{et al.} \cite{everythingatonce} designed a multi-modal fusion transformer that processes input from a combination of modalities. However, for text-to-video retrieval it claims that the fusion of video and audio modalities in their setup is not beneficial compared to only using visual information in a zero-shot setting. 
To the best of our knowledge, E\textsc{clip}SE \cite{ECLIPSE_ECCV22} is the first method to show the benefit of including audio in text-to-video retrieval in combination with strong pre-trained backbones. It introduces a symmetrical type of cross-attention for video and audio to align both modalities and shows its effectiveness in long-range video retrieval. In this work, we do not focus on the alignment between audio and video, but on the alignment between text-audio and text-video simultaneously. By extensive ablation studies, we show that it is not straightforward to combine the audio modality with other modalities for text-to-video retrieval, but we present a simple and effective text-conditioned feature alignment method to boost the performance by using audio.

\begin{figure*}
  \centering
    \includegraphics[width=0.92\linewidth]{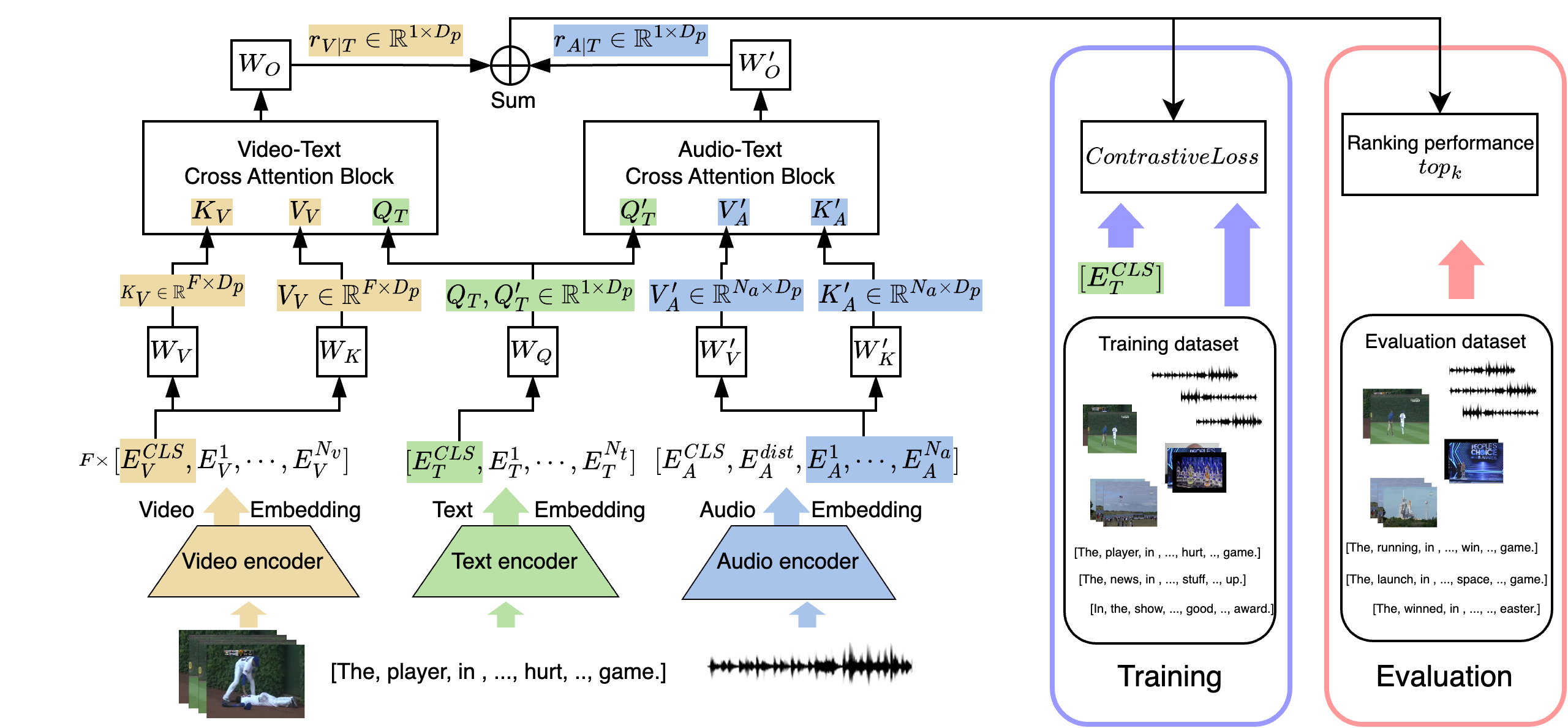}
    \caption{The model architecture of our method \textit{TEFAL} is presented in the left of the figure. The highlighted tokens from the encoders are the $F$ CLS tokens for video frames, one CLS token for the text caption and $N_{a}$ patch embeddings from the audio encoder. Query, Key and Value projections are created from these tokens and are used in the cross-attention blocks. These cross-attention blocks give us a text-conditioned video embedding and audio embeddings, which are fused through summation and used during training and evaluation, as presented on the right of the figure.}
  \label{fig:method} 
\end{figure*}

\section{Method}
In this section, we describe our proposed method \textit{TEFAL} to achieve \textbf{TE}xt-conditioned \textbf{F}eature \textbf{AL}ignment for audio-enhanced text-to-video retrieval. We evaluated multiple model architectures and the best results across several datasets were achieved by using two independent cross-modal attention blocks for the text to attend to the audio and video representations. The final representation of both the audio and video content are derived independently by conditioning on the text attention weights estimated in the corresponding cross-modal attention blocks. The complete model architecture, training setup and evaluation setup is presented in Figure \ref{fig:method}. We explain our key insights in \ref{subsec:method_key_insight_Text_feature_alignment}, followed by our overall architecture in \ref{subsec:method_overall_architecture}.

\subsection{Key Insight: Text-conditioned Feature Alignment}
\label{subsec:method_key_insight_Text_feature_alignment}
In most existing works, to achieve the multi-modality fusion, audio-video feature alignment is performed with direct feature fusion~\cite{wang2020alignnet} and more recently with cross-modal attention mechanisms~\cite{ECLIPSE_ECCV22}.

For the task of text-to-video retrieval, the goal is to align the text and video features in a joint latent space. However, the text is in general less expressive than the video and corresponds to a subset of the information provided by the video. Therefore, for text-to-video retrieval, the video representation would be better estimated conditioned on the text query to emphasize the aspects of the video which are more relevant to the text.

With these insights, our method, TEFAL, aligns the video frame tokens with text guidance in a text-video cross-attention block. Similar to \cite{gorti2022xpool}, the video frame embeddings ($\in \mathbb{R}^{F \times D_p}$) are the key and value inputs and the text embedding ($\in \mathbb{R}^{1 \times D_p}$) is the query input to the cross-attention computation, where $D_{p}$ is the embedding dimension of the respective features. Here, the cross-attention computation aims to condition the video frame tokens representation on the text query to obtain a weighted fusion based on the similarity between text and video frames. More importantly, the text tokens perform weighted token fusion in the frame dimension ($F$) to select the frames most similar to each of the text tokens.

Further, the audio modality in a video contains information key to identify the video itself. Thus, similar to the previously discussed cross-attention processing of video frames, our method aligns the audio tokens with the text query in a text-audio cross-attention block. Here, the audio embedding ($\in \mathbb{R}^{N_a \times D_p}$) is the key and value input and the same text embedding as above is the query input to the cross-attention computation. The cross-attention computation aims to weigh the tokens in the audio embedding to perform weighted fusion conditioned on the text representation. In the text-audio cross-attention, the representation of the audio tokens, corresponding to the input patches from the audio signal spectrogram, are updated conditioned on the text query.

\subsection{Overall Architecture}
\label{subsec:method_overall_architecture}
Inspired by the works of \cite{gorti2022xpool, Luo2022CLIP4ClipAE}, we bootstrap from joint image and text models. More specifically, we build on the CLIP \cite{clip} model as our text and video frame backbone and Audio Spectrogram Transformer (AST) \cite{gong21b_interspeech} as our audio backbone to obtain feature embeddings. For a single text-video pair, given a text description $T$, a video $V$ with $F$ video frames and an audio signal $A$, we compute their feature embeddings as follows:

\noindent \textbf{Video and Text Features} For each video frame $v^f$ we compute features for $N_v$ patches uniformly sampled in the spatial dimension. The text description for a video can vary in length and contains $N_t$ words. The CLIP model outputs a video frame embedding $\in \mathbb{R}^{(N_v +1) \times D}$ and a text embedding $\in \mathbb{R}^{(N_t+1)\times D}$, where the additional token is the class token $E^{CLS}_V$ and $E^{CLS}_T$ for video and text inputs, respectively. As the final video embedding $E_V$, we use $F$ CLS tokens $E^{CLS}_V$, one token for each video frame, and for text embedding $E_T$, one single text token $E^{CLS}_T$ that represents the query embedding.

\noindent \textbf{Audio Features} To compute audio features, the Mel-spectrogram features of the audio signal $A$ is passed as input to the encoder. We utilize the recent AST encoder model \cite{gong21b_interspeech} which demonstrated a strong performance in the audio classification task. From each spectrogram, $N_a$ patches are produced from adjacent overlapping windows to give a final set of $\in \mathbb{R}^{(N_a+2) \times D}$ tokens. For input audio embedding $E_A$ to the cross-attention component, we use $N_a$ patch embeddings and discard the two additional tokens, the CLS token ($E^{CLS}_A$) and the distillation token ($E^{dist}_A$).

\noindent \textbf{Text-conditioned Audio and Video Representations} 
The CLS token is well understood to have fused information from all other tokens~\cite{deit,gong21b_interspeech}. We utilize the CLS token as the final feature embedding for the video frame and the text input. The text-video cross-attention selects text-video frame similarity and takes the entire set of $F$ frames $\in \mathbb{R}^{F\times D}$ as input. Similar to the text-video cross-attention, the text-audio cross-attention selectively fuses audio patch embeddings based on the importance of the parts in the audio signal. Thus, we use all the $N_a$ tokens.

The final feature embeddings are projected into a common latent space, where the projections are defined as:
\begin{equation}
\begin{aligned}[c]
        Q_T &= \textrm{LN}(E_T^T)W_Q \\
        K_V &= \textrm{LN}(E_V)W_K \\
        V_V &= \textrm{LN}(E_V)W_V
\end{aligned}
\quad\leftrightarrow\quad
\begin{aligned}[c]
    Q'_T &= \textrm{LN}(E_T^T)W_Q'\\
    K'_A &= \textrm{LN}(E_A)W_K' \\
    V'_A &= \textrm{LN}(E_A)W_V'
\end{aligned}
\end{equation}
where the learned weight matrices $W_Q, W_K, W_V \in \mathbb{R}^{D \times D_p}$ and $W_Q', W_K', W_V'\in \mathbb{R}^{D \times D_p}$ project the final embeddings from $\mathbb{R}^{D}$ to $\mathbb{R}^{D_p}$. $\mathrm{LN}$ stands for LayerNorm.

The text query attends to the audio and video features via the scaled dot product attention ($\mathrm{XAttn}$), 
\begin{equation}
\begin{aligned}
     \mathrm{XAttn}(Q_T, K_V, V_V) & = \mathrm{softmax}\left(\frac{Q_T K_V^T}{\sqrt{D_p}}\right) V_V \\
     \mathrm{XAttn}(Q'_T, K'_A, V'_A) & = \mathrm{softmax}\left(\frac{Q'_T {K'_A}^T }{\sqrt{D_p}}\right) V'_A
    \end{aligned}
\end{equation}

In the text-video cross-attention, the text query attends to the per-frame video tokens in the key input and selectively fuses based on similarity between text and video tokens. Similarly, in text-audio cross-attention, the text query attends to the audio tokens from the full audio input and fuses based on similarity between text and audio tokens.

To get the final text conditioned audio and video embeddings, we project the cross-attention based output to the final output space via a weight matrix $W_O$.
\begin{equation}
\begin{aligned}
    E_{V|T} & =\textrm{LN}(\mathrm{XAttn}(Q_T, K_V, V_V)W_O) \\
    E_{A|T} & =\textrm{LN}(\mathrm{XAttn}(Q'_T, K'_A, V'_A)W'_O) 
    \end{aligned}
\end{equation}

The joint embedding $E_{(V,A)|T}$ for a single video is obtained by a simple addition of the text conditioned audio $E_{A|T}$ and video embedding $E_{V|T}$. In Section \ref{ablation-results} we will further elaborate on the effectiveness of this simple fusion method and make an experimental comparison with other potentially suitable fusion methods. 

\noindent \textbf{Text-to-Video Retrieval.} The final fused embedding $E_{(V,A)|T}$ is compared with the text query embedding $E_{T}$, which is the text CLS token $E^{CLS}_T$, via cosine similarity with the help of the following loss.

\noindent \textbf{Loss}. Our model is trained by using the infoNCE loss, a loss which is commonly used for contrastive learning since \cite{cpc} and later more specifically for training with image-text pairs such as in CLIP \cite{clip}. Consider having $K$ text and video-audio embedding pairs $\{(E_{T}^{i},E_{(V,A)|T}^{i}\}_{i=1}^{K}$. The infoNCE loss is applied on these pairs where a matching text caption and video are seen as a positive sample and all other caption-video combinations in the batch are seen as negatives. We optimize this loss in a symmetric way by using two losses, a text-to-video ($t2v$) and video-to-text ($v2t$) retrieval loss and taking the sum of these two as the total loss. 

\begin{equation}
    \mathcal{L}_{t2v} = -\frac{1}{B} \sum_{i=1}^{B}{\log\frac{\exp^{s(E_{T}^{i},E_{(V,A)|T}^{i}) \cdot \tau}}{\sum_{j=1}^{B}{\exp^{s(E_{T}^{i},E_{(V,A)|T}^{j}) \cdot \tau}}}} \\
    \label{eq:loss1}
\end{equation}
\begin{equation}
    \mathcal{L}_{v2t} = -\frac{1}{B} \sum_{i=1}^{B}{\log\frac{\exp^{s(E_{T}^{i},E_{(V,A)|T}^{i}) \cdot \tau}}{\sum_{j=1}^{B}{\exp^{s(E_{T}^{j},E_{(V,A)|T}^{i}) \cdot \tau}}}}\\
    \label{eq:loss2}
\end{equation}

\begin{equation}
    \mathcal{L} = \mathcal{L}_{t2v} + \mathcal{L}_{v2t}
    \label{eq:loss}
\end{equation}
where $s(E_{T}^{i},E_{(V,A)|T}^{j})$ is the cosine similarity between the text embedding $E_{T}^{i}$ and the fused audiovisual feature $E_{(V,A)|T}^{j}$, $B$ is the batch size and $\tau$ is a learnable scaling parameter, also known as the temperature. By bootstrapping from a pre-trained CLIP model and through our cross-modal attention mechanism, training with this loss enables our model to learn to match a text with its most semantically similar sub-regions of the ground-truth video. 

\section{Experiments}

We perform experiments on benchmark datasets for text-to-video retrieval that include audio tracks, and evaluate our performance following existing literature \cite{gorti2022xpool} and report the Recall@1 (R1), Recall@5 (R5), Recall@10 (R10), Median Rank (MdR), and Mean Rank (MnR) scores.

\subsection{Datasets}

\noindent{\textbf{MSR-VTT}} \cite{msrvtt} is considered as the most common dataset for text-to-video retrieval and the videos come with an audio track, consisting of 10,000 web video clips between 10-32 seconds. The dataset has two commonly used splits: one with 7k training videos and one with 9k, resulting in 140k and 180k video-caption pairs, respectively. Both splits use the same evaluation set of 1000 video-caption pairs. We report results on both splits.
Similar to \cite{hao2018integrating, everythingatonce, ECLIPSE_ECCV22}, we use the audio signals that are provided with the videos. 8,811 of the 10,000 videos have the audio track.

\noindent{\textbf{LSMDC}} contains 118,081 video clips from movies each paired with a single caption description. The lengths of videos range from 2 to 30 seconds. 101,079 videos are used for training, 7,408 for validation and 1,000 for testing, following the setting of X-Pool~\cite{gorti2022xpool}.

\noindent{\textbf{VATEX}} \cite{vatex} contains 34,991 video clips with multiple captions per video. We follow the HGR \cite{hgr} split protocol. There are 25,991 videos in the training set, 1,500 videos in the validation set and 1,500 videos in the test set. This dataset is regarded as one long-range video dataset.

\noindent{\textbf{Charades}} \cite{charades} contains 9,848 videos with one textual description per video. The average video length is 28 seconds. We follow \cite{ECLIPSE_ECCV22} in their train and test setup.

\subsubsection{Data Preprocessing} We leverage pre-trained backbones 
 as baselines to further improve our model's performance. The AST backbone is pre-trained on datasets where all audio files have the same duration. Correspondingly, both the frame shift $f_{shift}$ in the windowed Fourier transform and the target length $L_{tar}$ of the Mel Filter Bank (MFB) features are fixed. The CLIP video encoder, however, is pre-trained on datasets where videos have varying duration, and uniformly samples frames from videos. To reconcile the mismatch between the two preprocessings, we adaptively set the frame shift of MFB in audio preprocessing so that MFB features are uniformly sampled from the audio signal and that MFB feature length is fixed across samples. The frame shift used in filter bank calculation depends on the audio length. Specifically, the frame shift $f_{shift}$ in millisecond is calculated as $\mathbf{f_{shift}} = {n_{frm} \cdot 1000} / {(sr \cdot L_{tar})}$, where $n_{frm}$ is the number of sampled audio frames, $sr$ is the sampling rate in Hz and $L_{tar}$ is the target audio filter bank length.

\subsection{Implementation Details}
For the AST model, we use a data efficient image transformer DeiT \cite{deit} model, pre-trained on ImageNet and AudioSet \cite{audioset}. We finetune this on the audio files from the MSR-VTT dataset. Whenever a video does not have an audio file, we set the filter bank to a zero vector.

In our implementation, we set the embedding and projection dimensions as 512 ($D=D_p=512$).  We set the number of video frames $F$ to 12 for MSR-VTT and LSMDC and to 32 for VATEX and Charades, similar to X-Pool \cite{gorti2022xpool} and $\text{E\textsc{clip}SE}$. The final video input to the cross-attention module has 12 tokens of 512 dimensions. For the audio signal, we use audio sampling rate $sr=16k$ and set the target length $L_{tar}=1024$, which is the same as that used in ImageNet and AudioSet pretraining. The final patch embeddings that are input to the cross-attention block have 1212 tokens of 512 dimensions.

During finetuning we use a simplified strategy compared to the original AST \cite{gong21b_interspeech} and disable mixup and spectrogram masking. Our models with a ViT-B/32 backbone are trained with batch size 12 on a single A100 GPU and require 1 day of training time. Our models with a ViT-B/16 backbone are trained on 8 V100 GPUs with a batch size of 32.

\begin{table}
\centering
\resizebox{\linewidth}{!}{
\begin{tabular}{lc|ccccc}
\toprule
 Method & Modality  & R1 $\uparrow$ & R5 $\uparrow$ & R10 $\uparrow$ & MdR $\downarrow$ & MnR $\downarrow$ \\
 \midrule
X-Pool \cite{gorti2022xpool} & V  & 43.9 & 72.5 & 82.3 & 2.0 & 14.6 \\ 
X-Pool \cite{gorti2022xpool} & A &  5.6 &  14.7 & 21.3 & 109.0 & 157.4 \\ 
\midrule
\textbf{TEFAL} & A + V & \textbf{48.1} & \textbf{73.8} &  \textbf{82.8} & 2.0 & 12.1 \\ 
\bottomrule
 \end{tabular}}
 \caption{Results on MSR-VTT 7k that confirm the potential of audio as an additional signal for the text-to-video retrieval task. By using audio without the corresponding video, we show that a correspondence between text and audio can be learned. M denotes Modalities. V and A represent Video and Audio, respectively.}
  \label{tab:a+v}
\end{table}

\begin{table}[]
\centering
\resizebox{\linewidth}{!}{%
\begin{tabular}{l|ccccc}
\toprule
 Method  & R1 $\uparrow$ & R5 $\uparrow$ & R10 $\uparrow$ & MdR $\downarrow$ & MnR $\downarrow$ \\
 \midrule
Everything at once$\dagger$ \cite{everythingatonce}  & - & 62.7 & 75.0 & - & - \\
ALPRO* \cite{alpro} & 33.9 & 60.7 & 73.2 & 3.0 & -\\
$\text{CLIP4Clip}_{meanP}$ \cite{Luo2022CLIP4ClipAE}  & 42.1 & 71.9 & 81.4 & 2.0  & 16.2 \\
$\text{E\textsc{clip}SE}_{meanP}\dagger$ \cite{ECLIPSE_ECCV22}   & 43.2 & 71.5 & 81.9 & 2.0 & 15.9 \\
CenterCLIP \cite{centerclip} & 43.7 & 71.3 & 80.8 & 2 & 16.9 \\
X-Pool\cite{gorti2022xpool}  & 43.9 & 72.5 & 82.3 & 2.0 & 14.6 \\
\midrule
\textbf{TEFAL}  & \textbf{48.1} & \textbf{73.8} &  \textbf{82.8} & 2.0 & 12.1 \\
\bottomrule
 \end{tabular}}
 \caption{This table presents the results on MSR-VTT 7k. All models use ViT-B/32 backbones that are pre-trained on WebImageText. * indicates that the model is pre-trained on 5.5M additional text-image and text-video pairs
~\cite{clip}. $\dagger$ indicates that audio is used. - denotes that the value was not reported in the original paper.}
  \label{tab:7k}
\end{table}

\begin{table}[hbt!]
\centering
\resizebox{1\linewidth}{!}{%
\begin{tabular}{l|ccccc}
 \toprule
 Method  & R1 $\uparrow$ & R5 $\uparrow$ & R10 $\uparrow$ & MdR $\downarrow$ & MnR $\downarrow$ \\
 \midrule
 ViT-B/32 (backbone) \\
 \midrule
$\text{CLIP4Clip}_{meanP}$ \cite{Luo2022CLIP4ClipAE}   & 43.1 & 70.4 & 80.8 & 2.0 & 15.3 \\   
CenterCLIP \cite{centerclip} & 44.2 & 71.6 & 82.1 & 2 & 15.1 \\
$\text{E\textsc{clip}SE}_{meanP}\dagger$ \cite{ECLIPSE_ECCV22}   & 44.9 & 71.3 & 81.6 & 2.0 & 15.0 \\
BridgeFormer*\cite{bridgeformer}   & 44.9 & 71.9 & 80.3 & 2.0 & 15.3 \\     
X-CLIP\cite{xclip} & 46.1 & 73.0 & 83.1 & 2.0 & 13.2 \\
X-Pool\cite{gorti2022xpool}    & 46.9 & 72.8 & 82.2 & 2.0  & 14.3 \\
TS2-Net\cite{liu2022ts2net}  & 47.0 & 74.5 & 83.8 & 2.0 & 13.0 \\ 
CAMoE + DSL \cite{dsl} & 47.3 & 74.6 & 83.8 & 2.0 & 11.9 \\ 
\midrule
\textbf{TEFAL} & \textbf{49.4} & \textbf{75.9} & \textbf{83.9} & 2.0 & 12.0 \\
\midrule
\midrule
ViT-B/16 (backbone) \\
\midrule
OmniVL* \cite{wang2022omnivl} & 47.8 & 74.2 & 83.8 & - & - \\
\midrule
\textbf{TEFAL}  & 49.9 & 76.2 & 84.4 & 2.0 & 11.4 \\  
\textbf{TEFAL}+DSL  & 50.1 & \textbf{77.0} & 85.4 & 1.0 & 10.5 \\ 
\textbf{TEFAL}+DSL+QB-Norm & \textbf{52.0} & 76.6 & \textbf{86.1} & 1.0 & 11.4 \\     
 \bottomrule
 \end{tabular}}
 \caption{Results on MSR-VTT 9k split. All works use a CLIP ViT backbone which is pre-trained on this Wikipedia-based image-text dataset. Both ViT-B/32 and ViT-B/16 backbones are adopted for evaluation. Post-processing techniques, DSL \cite{dsl} and QB-Norm~\cite{qbnorm} are also used to boost the performance. * indicates the use of additional training pairs, more specifically BridgeFormer uses 5.5M additional training pairs and OmniVL uses 14M pairs.}
  \label{tab:9k}
\end{table}

\subsection{Experimental Results and Analysis}
\subsubsection{Main Results}
In Table~\ref{tab:a+v}, we present the results that support the motivation of our approach. First we evaluate on text-to-video and text-to-audio retrieval separately on MSR-VTT 7k. Although the results on text-to-audio retrieval are lower, R1 of 5.6\% compared to text-to-video retrieval result, R1 of 43.9\%, we see that combining the modalities for audio enhanced text-to-video retrieval gives us an improvement of 4\% on MSR-VTT 7k, compared to the text-to-video retrieval model only. 

In Tables~\ref{tab:7k} and \ref{tab:9k} we present the results on MSR-VTT 7k and 9k splits respectively and show that our model outperforms current state-of-the-art results. More specifically, for MSR-VTT 7k, our method is 4.9\% better in R1 than E\textsc{clip}SE \cite{ECLIPSE_ECCV22}, the current best method that uses audio for text-to-video retrieval. For MSR-VTT 9k we outperform E\textsc{clip}SE by 4.5\% for the R1. However, we also outperform other state-of-the-art methods with 2.1\%. This performance can be boosted even more with the use of a larger backbone (+0.5\%) and QB-Norm (+2\%). The supplementary material contains video-to-text retrieval and qualitative results.

In Table~\ref{tab:lsmdc}, \ref{tab:vatex}, and \ref{tab:charades}, we show that TEFAL also outperforms current state-of-the-art methods with an improvement of R1 of about 1\% for LSMDC and VATEX compared to the best previous method and up to 2.2\% for Charades. We can conclude that our methods can well deal with both short-range and long-range video datasets. 

\begin{table}
\centering
\resizebox{\linewidth}{!}{
\begin{tabular}{l|ccccc}
\toprule
 Method & R1 $\uparrow$ & R5 $\uparrow$ & R10 $\uparrow$ & MdR $\downarrow$ & MnR $\downarrow$ \\
 \midrule
 $\text{CLIP4Clip}_{meanP}$ \cite{Luo2022CLIP4ClipAE} & 20.7 & 38.9 & 47.2 & 13.0 & 65.3 \\
CenterCLIP \cite{centerclip} & 21.9 & 41.1 & 50.7 & 10.0 & 55.6 \\
TS2-Net\cite{liu2022ts2net} & 23.4 & 42.3. & 50.9 & 9.0 & 56.9 \\
X-Pool \cite{gorti2022xpool} & 25.2 & 43.7 & 53.5 & 8.0 & 53.2 \\
CAMoE + DSL \cite{dsl} & 25.9 & \textbf{46.1} & 53.7 & - & 54.4 \\
TEFAL & \textbf{26.8} & \textbf{46.1} & \textbf{56.5} & \textbf{7.0} & \textbf{44.4} \\
\bottomrule
 \end{tabular}}
 \caption{Results on the test split of LSMDC \cite{lsmdc}}
  \label{tab:lsmdc}
\end{table}

\begin{table}
\centering
\resizebox{\linewidth}{!}{
\begin{tabular}{l|ccccc}
\toprule
 Method & R1 $\uparrow$ & R5 $\uparrow$ & R10 $\uparrow$ & MdR $\downarrow$ & MnR $\downarrow$ \\
 \midrule
 $\text{CLIP4Clip}_{seqTransf}$ \cite{Luo2022CLIP4ClipAE} & 55.9 &  89.2 & 95.0 & 1.0 & 3.9\\
 $\text{E\textsc{clip}SE}_{meanP}\dagger$ \cite{ECLIPSE_ECCV22} & 57.8 & 88.4 & 94.3 & 1.0 & 4.3\\
TS2-Net\cite{liu2022ts2net} & 59.1 & 90.0 & 95.2 & 1.0 & \textbf{3.5} \\
X-Pool \cite{gorti2022xpool} & 60.0 & 90.0 & 95.0 & 1.0 & 3.8 \\
TEFAL & \textbf{61.0} & \textbf{90.4} & \textbf{95.3} & 1.0 & 3.8 \\
\bottomrule
 \end{tabular}}
 \caption{Results on the test split of VATEX \cite{vatex}}
  \label{tab:vatex}
\end{table}

\begin{table}
\centering
\resizebox{\linewidth}{!}{
\begin{tabular}{l|ccccc}
\toprule
 Method & R1 $\uparrow$ & R5 $\uparrow$ & R10 $\uparrow$ & MdR $\downarrow$ & MnR $\downarrow$ \\
 \midrule
 $\text{CLIP4Clip}_{meanP}$ \cite{Luo2022CLIP4ClipAE} & 13.9 & - & - & - & - \\
$\text{E\textsc{clip}SE}_{meanP}\dagger$ \cite{ECLIPSE_ECCV22}  & 15.7 & - & - & - & - \\
X-Pool \cite{gorti2022xpool} & 16.1 & 35.2 & 44.9 & 14.0 & 67.2 \\
TEFAL & \textbf{18.5} & \textbf{37.3} & \textbf{48.6} & \textbf{11.0} & \textbf{60.6} \\
\bottomrule
 \end{tabular}}
 \caption{Results on the test split of Charades \cite{charades}}
  \label{tab:charades}
\end{table}

\begin{table}
\centering
\resizebox{\linewidth}{!}{
\begin{tabular}{lcc|ccc}
\toprule
 Method & Finetuned AST & Adaptive $f_{shift}$  & R1 $\uparrow$ & R5 $\uparrow$ & R10 $\uparrow$  \\
 \midrule
TEFAL & $\checkmark$ & $\checkmark$  & \textbf{49.4} & \textbf{75.9} & \textbf{83.9} \\ 
TEFAL & $\checkmark$ &  & 48.6 & 74.7 & 84.1  \\
TEFAL &  & $\checkmark$ & 45.6 & 72.8 & 83.2  \\
\bottomrule
 \end{tabular}}
 \caption{A correct use of the audio encoder is crucial to achieve a good performance. Ablation study results show that finetuning the audio encoder and using an adaptive $f_{shift}$ based on the frame length give the best scores on MSR-VTT 9k.}
  \label{tab:feature}
\end{table}

\begin{table*}[hbt!]
\centering
\resizebox{0.72\linewidth}{!}{
\begin{tabular}{lll|ccc}
\toprule
 Method & Fusion Type & Expression & R1 $\uparrow$ & R5 $\uparrow$ & R10 $\uparrow$\\
 \midrule
TEFAL & Addition (best model) & $E_{(V,A)|T} = E_{V|T} + E_{A|T}$ & \textbf{49.4} & \textbf{75.9} & \textbf{83.9} \\ 
TEFAL & Late Fusion (X-Pool + audio) (A) & $E_{(V,A)|T} = E_{V|T} + E_{A}$ & 43.5 & 70.4 & 80.9  \\
TEFAL & Concatenation (B) & $E_{(V,A)|T} = \mathrm{FC}([E_{V|T},E_{A|T}])$ & 43.1 & 72.1 & 82.0  \\ 
TEFAL & Fusion by XAttn (C) & $E_{(V,A)|T} = \mathrm{XAttn}(E_{T},[E_{V|T},E_{A|T}])$ & 45.9 & 72.9 & 81.4  \\ 
TEFAL & Stacking audio and video (D) & $E_{(V,A)|T} = \mathrm{XAttn}(E_{T},[E_{V},E_{A}])$ & 46.1 & 71.8 & 81.8 \\ 
\bottomrule
\end{tabular}}
 \caption{An evaluation of alternative fusion types for the audio and video embeddings. The visual comparisons of these fusion methods are illustrated in Figure~\ref{fusion}. }
  \label{tab:fusion}
\end{table*}

\subsubsection{Audio Feature Extraction}
We do additional ablation study experiments on MSR-VTT 9k to evaluate the effects of different feature extraction methods for audio. The results are presented in Table~\ref{tab:feature}. In our setup we finetune the audio encoder, since we have seen in the second row of Table~\ref{tab:a+v} that finetuning the audio encoder in the text-to-audio retrieval setup actually helps for this task. We do the following ablation study experiments:

\noindent \textit{Freezing the AST}: in this setup we freeze the AST model and only train the last linear layer that reduces the embedding dimension from 768 to 512. This setup is similar to E\textsc{clip}SE \cite{ECLIPSE_ECCV22}, since they use the embeddings from the pre-trained audio encoders. We report an improvement of 3.8\% of R1 on finetuning vs freezing the AST model.
    
\noindent \textit{Fixing the frameshift}: a dynamic frame shift has been used in our method to make sure that the Mel Filter Bank maintains a fixed length and will not lose relevant information in case of long videos. We measure the effect of such dynamic frame shift value by fixing it to 10 seconds. The AST model uses a fixed frameshift of 10 seconds for all datasets as well as E\textsc{clip}SE \cite{ECLIPSE_ECCV22}. Taking a variable frameshift depending on the audio length improves the performance by 0.8\%.

\begin{figure*}
  \centering
    \includegraphics[width=0.76\linewidth]{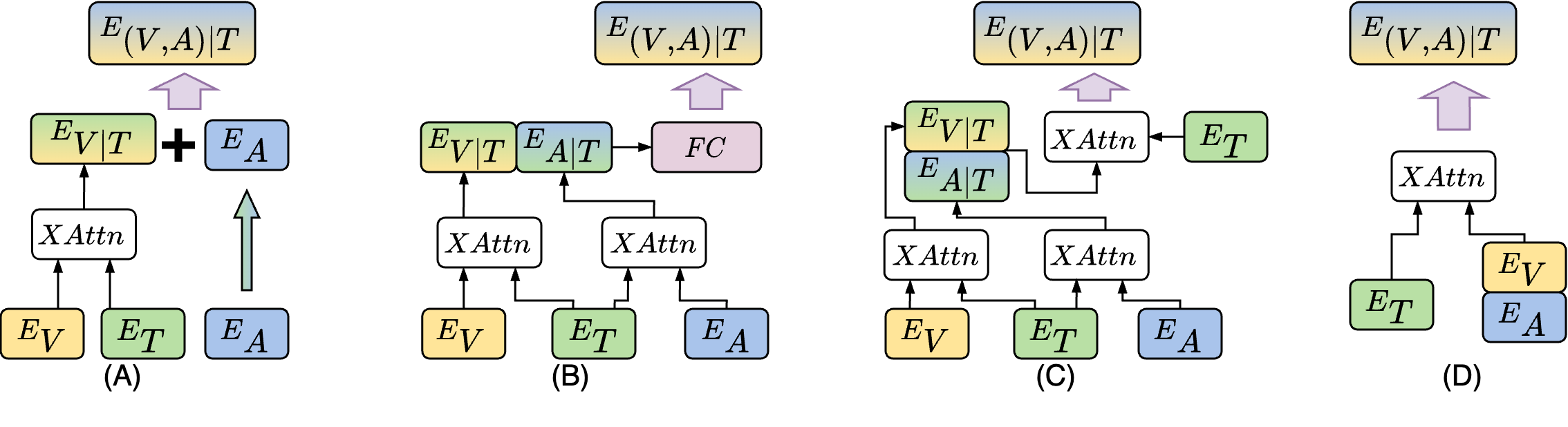}
  \caption{Different fusion methods corresponding to the results in Table \ref{tab:fusion} with (A) Late fusion, (B) fusion by concatenation in embedding dimension, (C) fusion by stacking the text-conditioned video and audio embeddings (D) fusion by stacking the video and audio embeddings.}
  \label{fusion}
\end{figure*}

\subsubsection{Variations on Multimodal Fusion}
\label{ablation-results}
We compare our fusion method, addition, with four other fusion types that are visualized in Figure~\ref{fusion}. Results are presented in Table~\ref{tab:fusion}, where $E_{V|T}$ stands for text-conditioned video embeddings (X-Pool) and $E_{A|T}$ stands for text-conditioned audio embeddings.

\noindent \textit{(A) Late fusion} : In this setup, instead of using text-conditioned audio embeddings, we use the audio embeddings from the audio encoder directly and fuse them with the text-conditioned video embeddings by using addition. 
    
    \begin{equation*}
    \small
    \textstyle
    E_{(V,A)|T} = E_{V|T} + E_{A}
    \end{equation*}
    
    This model shows a drop of 5.9\% in R1 compared to our best fusion technique, which shows the importance of explicit alignment between the text and audio embeddings.
    
\noindent \textit{(B) Concatenation} : Intuitive concatenation of the text-conditioned audio and text-conditioned video embeddings followed by a fully connected layer might help select the relevant components of each modality for each query. By concatenating the audio and video features, we get an overall embedding dimension of 1024, which is reduced to 512 with a fully connected layer to allow the computation of cosine similarity with each text embedding. But we noticed a large drop in performance of 6.3\% in R1 using this method.
    \begin{equation*}
    \small
    \textstyle
    E_{(V,A)|T} = \mathrm{FC}([E_{V|T},E_{A|T}])
    \end{equation*}
    
    We argue that the linear layer at this stage of the model is possibly causing too many additional parameters to learn an efficient embedding.
    
\noindent \textit{(C) Fusion by cross-attention} : In this setup, we stack the text-conditioned audio and video embeddings and apply a third cross-attention block which takes the text embedding again as the query and the stacked audio-video embedding as key and value. This results in 
    \begin{equation*}
    \small
    \textstyle 
    E_{(V,A)|T} = E_{(V,A)|T} = \mathrm{XAttn}(E_{T},[E_{V|T},E_{A|T}])
    \end{equation*}
    
    The R1 score is 3.5\% lower than for the best model, which indicates that a third cross-attention block in its current design is not able to capture importance of the audio and video modalities related to the text better than addition.
    
    \noindent \textit{(D) Stacking audio and video} : The cross-attention block in TEFAL learns the weight related to the importance of each frame to the text. In this experiment, instead of using all audio patch embeddings, we use the average of the CLS and DIST tokens from the DEiT model and use this in one cross-attention block together with the video frame embedding. We could see the audio embedding as an additional frame embedding, by stacking the embeddings of the two modalities in the dimension of the number of the frames. 
    \begin{equation*}
    \small
    \textstyle
    E_{(V,A)|T} = \mathrm{XAttn}(E_{T}, [E_{V}, E_{A}])
    \end{equation*}

This leads to an R1 of 46.1\%. Our intuition is that the initial embedding spaces of the audio and video frame embeddings are unaligned and therefore require two cross-attention blocks to align them with the text embedding.

From the experiments regarding the fusion of the audio and video embeddings, we can conclude that fusion is not trivial. In fact, alternative fusion methods show lower results than X-Pool, the method that only uses the same video branch as TEFAL and does not leverage audio. The proposed simple addition fusion obtains an R1 of 49.4\%, which is the best among all the fusion methods and improves the result by 2.5\% compared to the video-only branch.

\subsubsection{Applicability at Scale} 
Our method computes aggregated video and audio embeddings that are both conditioned on text. Therefore, we cannot pre-compute the final audio and video embeddings offline, since the text queries are not known at this point. Similar to X-Pool, we can use TEFAL to re-rank the top $\mathcal{K}$ retrievals of an efficient method, i.e., mean-pooling the frame embeddings. Given $\mathcal{T}$ text queries, $\mathcal{V}$ videos and $\mathcal{A}$ audios, with $\mathcal{A} \le \mathcal{V}$, instead of a complexity of $\mathcal{O(TV)}$ we get $\mathcal{O(KT + V)}$, which is the same as X-Pool \cite{gorti2022xpool}. To empirically show that this does not result in a significant performance drop, we reduce the search space via approximate nearest neighbor method between the text embedding and the mean-pooled video embeddings, and then perform re-ranking in this reduced search space with TEFAL. Since all test sets have 1,000 to 2,000 videos, we also evaluate on the validation set of LSMDC, which has 7,408 videos, to show the effect of scaling up. In table \ref{tab:scale} we notice a very small drop in performance by using TEFAL to re-rank the top 10\% retrievals given by mean-pooling, albeit the latter model's low performance. 

\begin{table}[h]
\centering
\resizebox{\linewidth}{!}{
\begin{tabular}{l|ccc|ccc|ccc}
\toprule
\multicolumn{1}{c}{} & \multicolumn{3}{c}{mean-pool} & \multicolumn{3}{c}{TEFAL} & \multicolumn{3}{c}{TEFAL (re-rank)}\\
\cmidrule(rl){2-4} \cmidrule(rl){5-7} \cmidrule(rl){8-10}
Dataset & {R1} $\uparrow$ & {R5} $\uparrow$ & {R10} $\uparrow$ & {R1} $\uparrow$ & {R5} $\uparrow$ & {R10} $\uparrow$ & {R1} $\uparrow$ & {R5} $\uparrow$ & {R10} $\uparrow$ \\
\midrule
MSR-VTT-9k & 41.7 & 69.7 & 78.3 & 49.4 & 75.9 & 83.9 & 49.4 & 75.8 & 83.8 \\
LSMDC (test) &  20.9 & 38.5 & 48.3 & 26.8 & 46.1 & 56.5 & 26.7 & 46.1 & 56.2 \\
LSMDC (val) & 7.8 & 18.9 & 25.3 & 10.2 & 23.6 & 31.1 & 10.2 & 23.6 & 31.3 \\
VATEX &  53.3 &  84.9 & 92.2 & 61.0 &  90.4 & 95.3 & 61.0 & 90.3 & 95.2 \\
Charades & 11.8 & 28.0 & 36.8 & 18.5 & 37.3 & 48.6 & 18.6 & 37.7 & 49.3 \\
\bottomrule
\end{tabular}}
\caption{Results on all datasets showing the effect of a reduced search (by 90\%) space during inference.}
  \label{tab:scale}
\end{table}

\section{Conclusion}
This paper introduces TEFAL, a novel text-conditioned feature alignment framework for audio-enhanced text-to-video retrieval. Our approach utilises two independent cross-modal attention blocks for the text to attend to the audio and video representations. We are the first to propose this approach for audio-enhanced text-to-video retrieval and explain why it is more suitable for this task than audiovisual alignment.  Extensive experiments demonstrate that the text-conditioned feature alignment outperforms audiovisual alignment for audio-enhanced text-to-video retrieval. We attribute this success to our use of an independent cross-modality attention model that develops a representation conditioned on the text and independent of the video content. In the future, we plan to extend our method to other multimodal text-video-audio understanding tasks, such as video captioning and video question answering.

{\small
\bibliographystyle{ieee_fullname}
\bibliography{egbib}

}

\clearpage

\ificcvfinal
\thispagestyle{empty}
\fi
\appendix

\setcounter{table}{0}
\setcounter{figure}{0}
\renewcommand{\thetable}{T\arabic{table}}
\renewcommand{\thefigure}{F\arabic{figure}}

\section{Video-to-Text Retrieval Results}

In our work, we focus on the task of text-to-video retrieval. We follow other works by also evaluating our model trained on video-to-text retrieval for MSR-VTT 9k, since competing methods have only provided video-to-text retrieval results on this data. 

\begin{table}[hbt!]
 \centering
 \resizebox{1\linewidth}{!}{%
 \begin{tabular}{l|ccccc}
  \toprule
  Method  & R1 $\uparrow$ & R5 $\uparrow$ & R10 $\uparrow$ & MdR $\downarrow$ & MnR $\downarrow$ \\
  \midrule
 $\text{CLIP4Clip}_{meanP}$ \cite{Luo2022CLIP4ClipAE}   & 43.1 & 70.5 & 81.2 & 2.0 & 12.4 \\  
  X-Pool\cite{gorti2022xpool}    & 44.4 & 73.3 & 84.0 & 2.0 & 9.0 \\
 $\text{E\textsc{clip}SE}_{meanP}\dagger$ \cite{ECLIPSE_ECCV22}   & 44.7 & 71.3 & 82.8 & 2.0 & 10.8 \\
 BridgeFormer*\cite{bridgeformer}   & 44.9 & 71.9 & 80.3 & 2.0 & 15.3 \\     
CAMoE \cite{dsl} & 45.1 & 72.4 & 83.1 & 2.0 & 10.0 \\ 
 TS2-Net \cite{liu2022ts2net}  & 45.3 & 74.1 & 83.7 & 2.0 & 9.2 \\ 
 X-CLIP \cite{xclip} & 46.8 & 73.3 & 84.0 & 2.0 & 9.1 \\
 \midrule
 \textbf{TEFAL} & \textbf{47.1} & \textbf{75.1} & \textbf{84.9} & 2.0 & \textbf{7.4} \\
  \bottomrule
  \end{tabular}}
 \caption{Video-to-Text Retrieval Results on MSR-VTT 9k split. All works use a CLIP ViT-B/32 backbone which is pre-trained on this Wikipedia-based image-text dataset. }
   \label{tab:v-t-9k}
 \end{table}

\section{Qualitative Results}
\begin{figure*}[t]
\includegraphics[width=0.95\linewidth]{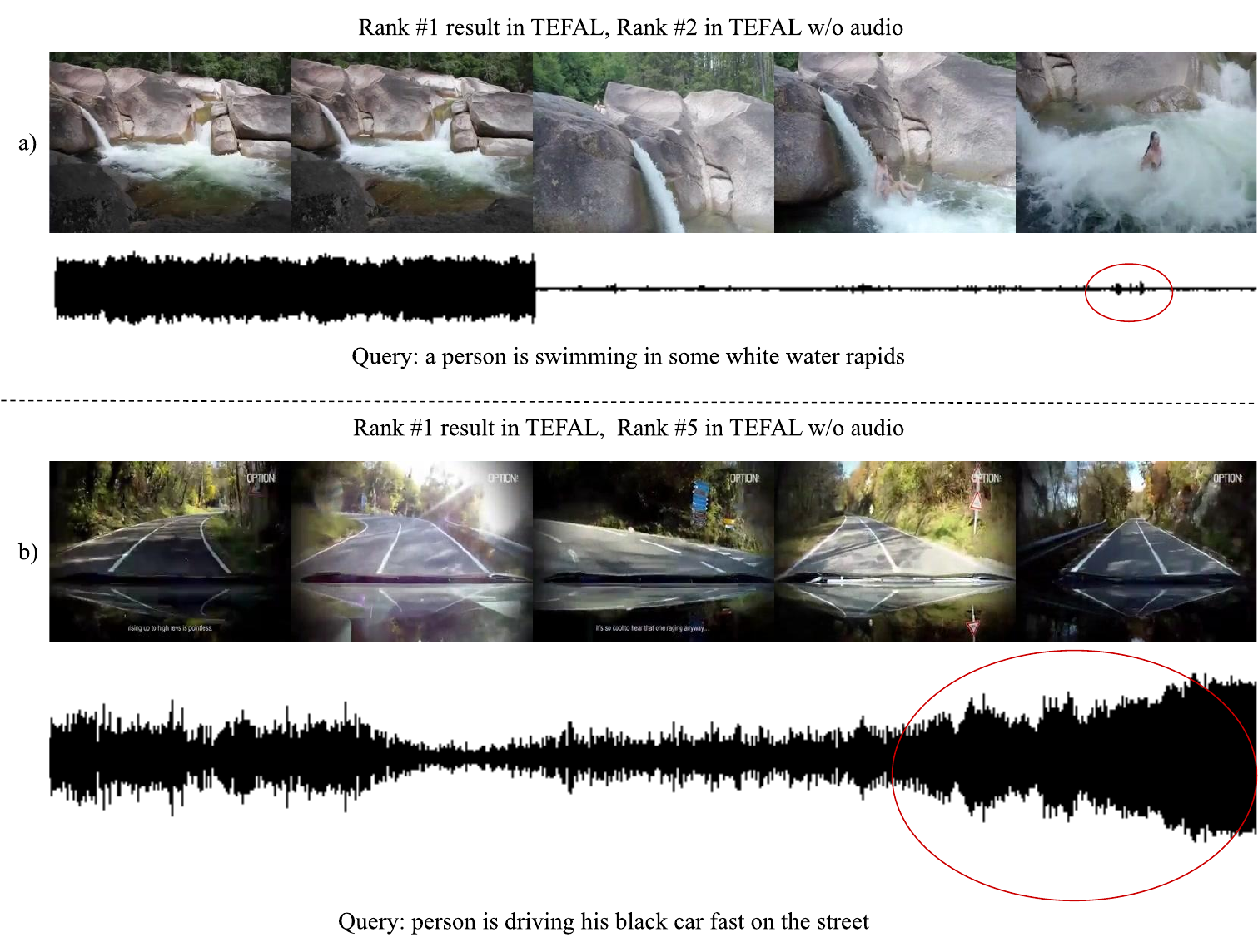}
\caption{In Figure a) an example is presented where a small sound has a large contribution to the final result. While TEFAL w/o audio is not able to select the correct video, TEFAL uses the audio to select the correct video as Rank 1. In Figure b) an example is shown of a description in the text that is not visible, namely the car accelerating, but is audible. Therefore the TEFAL model is able to perform much better on this (Rank 1) than the TEFAL w/o audio model (Rank 5)}
\label{fig:supp1}
\end{figure*}

In this section, we present specific examples on the MSR-VTT dataset \cite{msrvtt} to highlight how audio provides complementary information to the video to achieve improved text-queried retrieval. Additionally, these examples further justify our choice of using text features as the center feature to align video and audio features rather than aligning the audio and video modalities. Shown in Figure \ref{fig:supp1} are two examples where TEFAL ranks the matched video as the top retrieval with the help of audio modality, whereas, X-Pool \cite{gorti2022xpool} which only utilizes text and video fails to do so. (Note that TEFAL w/o audio corresponds to X-Pool that does not have an audio branch). 

\textbf{Example 1a.} The query words of the upper example (sample 7152) is ``a person is swimming in some white water rapids". While the video modality alone shows both the rapid water and the person, TEFAL w/o audio (video-only model) ranks the clip as the second matched retrieval. TEFAL, with the addition of the audio cue, correctly ranks the matched clip as the top retrieval. We notice that the presence of a person is confirmed by the voice in the latter part of the waveform (encircled in red), which clearly demonstrates that our model picks up complementary information from the audio modality. It is also observed that the first part of the clip is dominated by loud sound of streaming water but the water sound is greatly suppressed later in the clip though the continuous presence of water flowing in the video. This explicitly justifies building independent text-video and text-audio cross-modal attention blocks rather than aligning video and audio embeddings as it is in E\textsc{clip}SE \cite{ECLIPSE_ECCV22}, since the mandatory alignment between video and audio may introduce additional noise in the audiovisual feature.

\textbf{Example 1b.} The query words of the lower example (sample 9806) is ``person is driving his black car fast on the street". Similarly, the TEFAL w/o audio model ranks the matched video at the fifth place while TEFAL ranks the matched video as the top retrieval. In this example, the keyword ``fast" is crucial. However, driving fast is not visible in the frames alone but is cued by the sound of the accelerating engine (encircled in red in the waveform). 

\begin{figure*}[t]
\centering
   \includegraphics[width=0.7\linewidth]{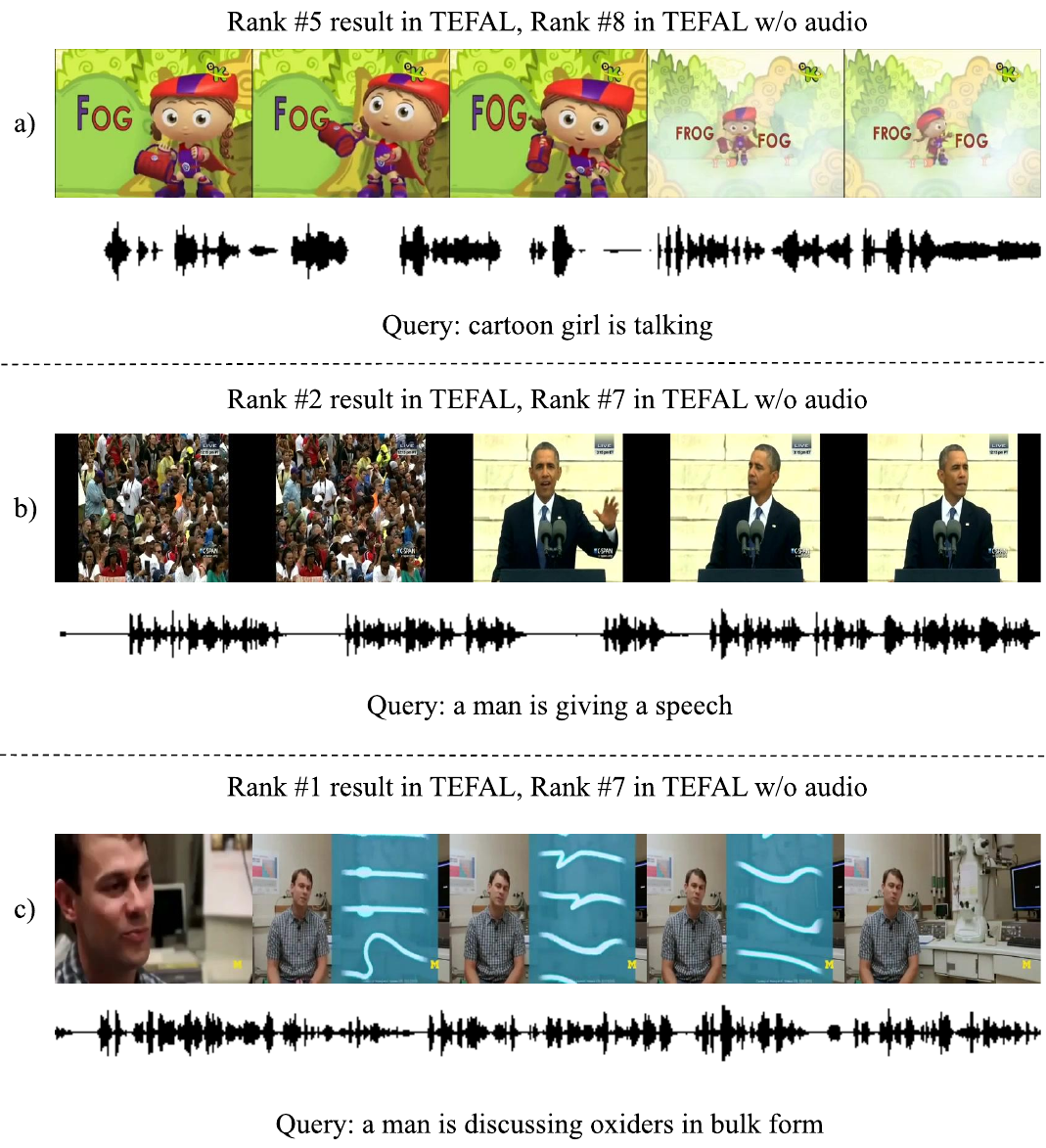}
     \caption{This Figure shows three examples that illustrate the correspondence between the text and audio modality, that contains the verb ``speaking'', ``talking'' or a variation and specific words that correspond to the text query.}
     \label{fig:supp2}
\end{figure*}

Additional examples are shown in Figure \ref{fig:supp2} to illustrate the correspondence between the text and audio modality that is otherwise missed between text and video. In the upper example, the girl talking can only be heard in the audio signal (sample 8827); in the middle example (sample 9233), the speech content is explicitly presented in the audio rather than video; and in the lower example, the word ``oxiders" can only be matched in the audio from the man's talking (sample 9249). 

\section{Visualization of Attention}
\begin{figure*}[t]
\centering
   \includegraphics[width=0.7\linewidth]{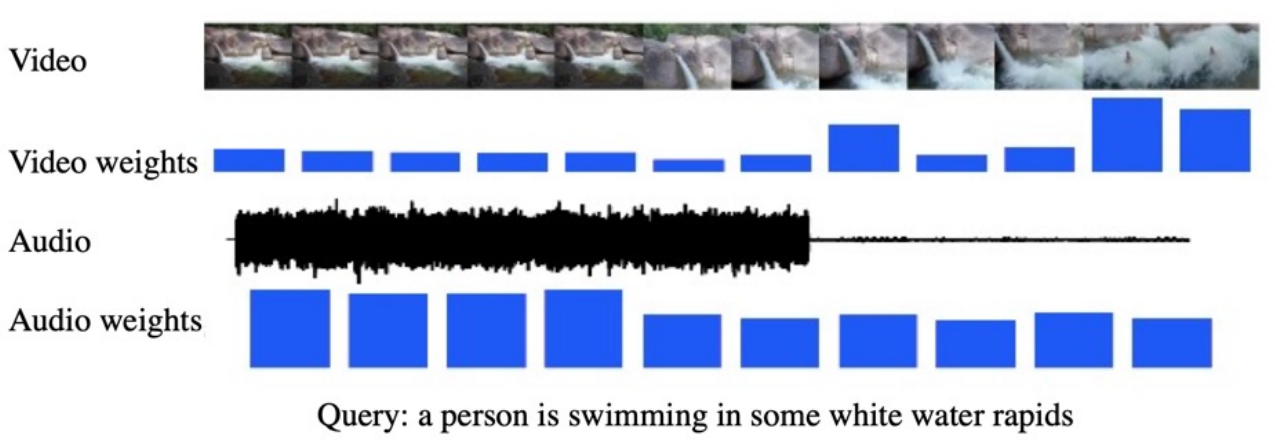}
     \caption{This figure shows the weights of frames from the text-video attention block (upper two rows) and the weights of audio tokens from the text-audio block (lower two rows). Video and audio weights emphasize complementary parts of the video. }
     \label{fig:supp3}
\end{figure*}

We provide a qualitative example (sample 7152) to illustrate the activated video and audio regions by plotting the attention weights in the $E_{V|T}$ and $E_{A|T}$ blocks in Figure \ref{fig:supp3}. We observe that the latter half of the video stream has higher activation than the first half, whereas, the first half of the audio has higher activation than the latter half, providing complementary information.

\section{Limitations}
The main limitation of TEFAL is that without audio the method reduces to the text-video branch, and the performance is similar with XPool \cite{gorti2022xpool} (as indicated by Figure 1 in the main manuscript). If the missing audios are mostly in the train set, meta learning approaches could help lessen this issue \cite{ma2021smil}.

\end{document}